\documentclass[sigplan,screen]{acmart}
\AtBeginDocument{%
  \providecommand\BibTeX{{%
    \normalfont B\kern-0.5em{\scshape i\kern-0.25em b}\kern-0.8em\TeX}}}

\begin{document}

\title{Evaluating the Performance of Large Language Models in Scientific Claim Detection and Classification}

\author{Tanjim Bin Faruk}
\email{tanjim@colostate.edu}
\affiliation{%
  \institution{Colorado State University}
  \city{Fort Collins}
  \state{Colorado}
  \country{USA}
}


\begin{abstract}

The pervasive influence of social media during the COVID-19 pandemic has been a double-edged sword, enhancing communication while simultaneously propagating misinformation. This \textit{Digital Infodemic} has highlighted the urgent need for automated tools capable of discerning and disseminating factual content. This study evaluates the efficacy of Large Language Models (LLMs) as innovative solutions for mitigating misinformation on platforms like Twitter. LLMs, such as OpenAI's GPT and Meta's LLaMA, offer a pre-trained, adaptable approach that bypasses the extensive training and overfitting issues associated with traditional machine learning models. We assess the performance of LLMs in detecting and classifying COVID-19-related scientific claims, thus facilitating informed decision-making. Our findings indicate that LLMs have significant potential as automated fact-checking tools, though research in this domain is nascent and further exploration is required. We present a comparative analysis of LLMs' performance using a specialized dataset and propose a framework for their application in public health communication.

\end{abstract}


\keywords{Large Language Models, Misinformation, Scientific Claim, Twitter Dataset}


\maketitle

\section{Introduction}


Over the past ten years, social media platforms have fundamentally altered the dynamics of interpersonal communication and public engagement. This shift became particularly pronounced during the COVID-19 pandemic when stay-at-home mandates relegated digital platforms such as Twitter to a central role in disseminating information. These platforms, while facilitating connectivity, also became the primary conduit for news and updates regarding health advisories and the progression of the pandemic. However, with the increased volume of user-generated content, the distinction between accurate information and misinformation became blurred, resulting in the proliferation of unsubstantiated rumors and fallacies about the virus on Twitter. The dissemination of such misinformation, particularly when propagated by influential entities on the platform, hindered the public's ability to make well-informed decisions. Given the overwhelming surge of data—termed the "Digital Infodemic"—the burden of verifying the veracity of COVID-19-related information became impractical for the general populace. This scenario underscored the imperative for an automated mechanism to manage the deluge of information associated with COVID-19. In this context, Large Language Models (LLMs) emerge as a potent tool to aid in the mitigation of misinformation spread on digital platforms.

Large Language Models (LLMs) represent a significant breakthrough in artificial intelligence, harnessing vast amounts of data to understand and generate human language with remarkable accuracy. These sophisticated models, epitomized by systems such as \textit{GPT} (Generative Pretarained Transformer) by OpenAI and \textit{LLaMA} (Large Language Model Meta AI) by Meta, are trained on diverse internet text, enabling them to perform a wide array of language tasks. They can generate coherent and contextually relevant text, answer questions, summarize lengthy documents, translate languages, and even create content that resonates with human readers. 

LLMs function by predicting subsequent units of text based on the input they receive, a capability that stems from their intricate neural network architectures and extensive parameter sets. The breadth of training received by LLMs endows them with a nuanced understanding of language, which transcends the capabilities of traditional machine-learning models. Traditional approaches necessitate laborious and time-consuming processes to curate a supervised model, including assembling a task-specific dataset, training the model, and optimizing it to yield reliable output—a procedure that could span several days. Moreover, such models are often susceptible to overfitting and exhibit limited adaptability beyond their initial training scope.

Conversely, LLMs streamline the workflow significantly. Given their pre-trained nature, LLMs obviate the need for prior task-specific training, enabling the deployment of a functional solution within hours. Their inherent design allows them to generalize across tasks, rendering them versatile for an array of applications beyond the confines of a single domain. Consequently, agility and efficiency position LLMs as invaluable assets in addressing challenges such as the misinformation crisis observed on social media platforms during the COVID-19 pandemic. By automating the detection and analysis of information, LLMs can assist in distinguishing credible sources from unreliable ones, thus supporting the dissemination of factual content and aiding the public in navigating the overwhelming tide of digital information.

Until now, the focus for misinformation detection has primarily been on utilizing pre-trained language models like SciBERT and RoBERTa, as highlighted in studies \cite{MDEFND} \cite{Kaliyar2021} \cite{zhu2022} \cite{mosallanezhad2022}. Recent research, such as that by \cite{hu2023bad} and \cite{huang2023harnessing}, has shifted towards evaluating the efficacy of GPT-3.5 in identifying text-based misinformation. However, comprehensive studies assessing the performance of various LLMs, specifically in the context of misinformation detection, are currently lacking. Additionally, the application of LLMs in detecting COVID-19-related misinformation, using specialized datasets, remains an area with limited exploration.


This study focuses on evaluating the performance of Large Language Models in detecting and classifying scientific claims related to COVID-19 topics. The analysis of scientific claims can help understand the spread of accurate information or misinformation \cite{lbekv} and automating the process can help us curb the spread of misinformation.

In summary, this study makes the following contributions:

\begin{itemize}
    \item Provides an in-depth analysis of the capacity of LLMs, such as GPT and LLaMA, to accurately identify and classify scientific claims related to COVID-19, an area with limited prior research focus.
    \item Evaluates the performance of LLMs using a specialized dataset, contributing novel insights into their role in combating misinformation during critical global health crises.
    \item Presents a potential framework for leveraging LLMs as automated fact-checking tools, addressing the pressing need for effective strategies to manage the "Digital Infodemic" and enhance public health communications.
\end{itemize}


The remainder of this paper is structured as follows:: Section \ref{related-work} presents a review of the literature relevant to this subject. Section \ref{background-information} delineates the background information necessary for understanding the context of the study. Section \ref{methodology} elaborates on the methodology employed. Section \ref{setup} outlines the experimental setup. Section \ref{evaluation} describes the evaluation metrics utilized in this research. Sections \ref{results} and \ref{result-analysis} detail the principal findings of the study. Section \ref{future-work} proposes potential directions for future research within this project. Finally, Section \ref{conclusion} summarizes the study, discussing its current limitations and suggesting avenues for future inquiry.

\section{Related Work} \label{related-work}

Initial studies in misinformation detection predominantly employed pre-trained language models like SciBERT and RoBERTa. Works such as those by \cite{Kaliyar2021} and \cite{zhu2022} utilized these models to detect misinformation by leveraging their language understanding capabilities. However, these studies often did not account for the specialized nuances of pandemic-related misinformation.

\cite{cdibtp} used embedding-based classifiers and BERT-based transfer learning to detect claims in biomedical Twitter posts. While this approach demonstrated potential, it was limited by the scope of BERT's pre-training and its capacity for domain-specific adaptation.

The study by \cite{sciadv.adh1850} showcased how models like GPT-3 could be harnessed for generating accurate information and assisting in fact-checking. Although promising, this research did not fully explore the model's performance across a diverse set of LLMs or in the context of COVID-19 misinformation.

\cite{wadden-etal-2022-scifact} introduced SciFact-Open, a dataset to evaluate scientific claim verification systems. Similarly, \cite{saakyan2021covidfact} developed COVIDFact, a dataset with claims concerning the COVID-19 pandemic. While these datasets are crucial for developing and testing misinformation detection systems, they often do not leverage the comprehensive capabilities of LLMs.

\cite{hafid2022scitweets} developed a framework for classifying tweets into various categories, which aligns closely with the objectives of our study. However, it does not explore the application of LLMs in this domain.

Recent studies by \cite{hu2023bad} and \cite{huang2023harnessing} focused on the capabilities of LLMs like GPT-3.5 in detecting misinformation. These studies mark a shift towards utilizing the advanced processing power of LLMs but do not provide a comparative analysis across multiple LLMs or specifically address the detection of COVID-19-related misinformation using specialized datasets.

The current literature provides a foundation for employing LLMs in misinformation detection. However, there is a discernible gap in comprehensive analyses of LLMs' performance, particularly in the unique context of the COVID-19 pandemic. This study aims to address this by evaluating the performance of multiple LLMs on a specialized dataset, providing insights into their effectiveness in mitigating the spread of misinformation during critical health crises.

\section{Background Information} \label{background-information}

\subsection{Scientific Claim Framework}

This research is particularly concerned with examining assertions pertaining to COVID-19, including topics related to the virus's nature, its origins, mask-wearing protocols, quarantine measures, and more. Central to this endeavor is the capability to discern the presence of claims within tweets and to substantiate such claims. The framework presented by \cite{hafid2022scitweets} offers a foundational approach for achieving this goal. The annotation methodology developed by these researchers includes specific directives for ascertaining the existence and confirmability of claims. By integrating these directives, Large Language Models (LLMs) can be equipped to intelligently analyze the content of tweets and determine both the presence and the potential for verification of claims. The ability to identify and validate scientific claims in tweets via scientific research enables the differentiation between authentic and fabricated content.

\subsection{Scientific Claim}

A scientific claim is defined by \cite{hafid2022scitweets} as one that is amenable to validation using scientific techniques or one for which evidence exists within the scientific literature. A text can encompass multiple claims or queries. Such a text is deemed scientifically verifiable provided that at least one constituent claim or inquiry meets the criteria for scientific verifiability.

\subsection{Scientific Claim Existence \& Verifiability}

Initially, it must be ascertained whether the content of a tweet pertains specifically to COVID-19-related subjects. Upon confirming its relevance, and if it contains assertions regarding COVID-19, subsequent analysis is conducted to determine the scientific verifiability of those claims.

\cite{hafid2022scitweets} offers the following criteria to identify a verifiable claim:

\begin{itemize}
    \item The claim is corroborated by scientific studies or literature.
    \item The claim is theoretically capable of being verified through scientific methods.
    \item The claim presents difficulties in verification.
    \item The claim is characterized by absurdity, humor, or irony.
\end{itemize}

Conversely, certain conditions suggest that a tweet's content may not be amenable to scientific verification:

\begin{itemize}
    \item The claim lacks support from scientific research literature.
    \item The content is reflective of an individual's personal views on matters unrelated to COVID-19.
\end{itemize}

\subsection{Examples and Counterexamples in Claim Classification}

The following examples and counterexamples illustrate the claim classification process:

\begin{itemize}
    \item \textbf{Example Tweet 1}: “I don’t think face masks and vaccines are effective against the COVID-19 virus.”
    \begin{itemize}
        \item Initially, we identify the implicit claim, "Face masks and vaccines are not effective against COVID-19 virus."
        \item Next, we assess whether this claim can be verified through scientific methods. If verifiable, the claim is categorized as "scientifically verifiable" and in this case, it is \cite{covid-nhs-1}.
    \end{itemize}

    \item \textbf{Example Tweet 2}: “We continue to remind everyone to follow the health protocols of wearing face masks and shields, safe physical distancing, and proper hygiene, among others, to ensure our safety from COVID-19.”
    \begin{itemize}
        \item Here, the claim "Wearing face masks and physical distancing ensures safety from COVID-19" is identified.
        \item This claim can also be checked against scientific research \cite{PRAKASH2022101091} to determine its verifiability.
    \end{itemize}

    \item \textbf{Counterexample Tweet}: “COVID-19 has slowed down restaurant business.”

    \begin{itemize}
        \item While this tweet does contain a claim, it focuses on the economic impact of COVID-19 ("slowed down restaurant business") rather than directly addressing COVID-19 health topics as specified earlier.
        \item Therefore, this tweet wouldn't be classified under "scientifically verifiable claims" related to COVID-19 health topics.
    \end{itemize}
\end{itemize}

These examples demonstrate the approach of identifying claims within tweets and evaluating their scientific verifiability. The process involves discerning the underlying assertion in each tweet and then determining its substantiality through scientific research, distinguishing between claims directly related to COVID-19 health topics and those that are not.

\section{Methodology} \label{methodology}

\subsection{Solution Sketch}

Our approach involves processing individual tweets from a previously compiled and manually annotated dataset \cite{covid19dataset}. Each tweet is fed into the Large Language Model (LLM) accompanied by a system prompt and various configuration settings. Utilizing techniques such as few-shot prompting and instructional patterns, the LLM assesses the tweet’s content. Guided by the criteria outlined in the system prompt, the model determines the existence of a claim within the tweet and evaluates its scientific verifiability. The model’s response, formatted as specified in the system prompt, includes the rationale behind its assessment.

\subsection{Dataset}

The dataset employed in this study \cite{covid19dataset} comprises $1847$ tweets, gathered through the Twitter Streaming API using keyword filtering to focus on significant tweets related to COVID-19. The selection process was informed by a dynamically updated list of pandemic-related keywords, such as \textit{coronavirus}, \textit{quarantine}, and \textit{face masks}.

\subsubsection{Dataset Description}

The dataset is organized into three distinct columns:

\begin{itemize}
    \item \textbf{Polished Text}: This column includes the content of the tweets. During preprocessing, all special characters were removed, and emojis were converted into their corresponding textual descriptions.
    \item \textbf{Claim}: This column reflects the manual annotation indicating the presence of a scientific claim in each tweet. It is represented by binary values, where each tweet is labeled either $0$ (no claim) or $1$ (claim present).
    \item \textbf{cat1}: This column contains manual annotations regarding the scientific verifiability of each tweet. A value of $1$ in this column corresponds to tweets where the \textit{Claim} column is also $1$, signifying verifiable content. In cases where the \textit{Claim} column is $0$, this column will also be $0$, indicating non-verifiability.
\end{itemize}

This structure ensures a clear and systematic approach to categorizing tweets based on their content and the presence of scientifically verifiable claims.

\subsubsection{Manual Data Annotation}

Each tweet in the data set was annotated by two annotators and was arbitrated by a third annotator in case of disagreements. The annotators assigned binary indicators to signify the presence or absence of specific scientific content types.

\subsubsection{Data Distribution}

Table \ref{dataset-statistics} shows the distribution of scientific claim existence and verifiability in the dataset.

\begin{table}[htbp]
\centering
\setlength{\abovecaptionskip}{-15pt} 
\setlength{\belowcaptionskip}{10pt} 
\begin{tabular}{@{}lcc@{}}
\toprule
\textbf{Task} & \textbf{Label} & \textbf{Total (\%)} \\
\midrule
Scientific Claim & Claim Existence & 922 (49.6\%)  \\
                 & Verifiable      & 729 (39.2\%)  \\
\bottomrule
\end{tabular}
\caption{Overview of Labels in the Data Set}
\label{dataset-statistics}
\end{table}

\subsection{Selection of Models}

The decision regarding which models to employ encompassed both proprietary and open-source options. In the proprietary category, GPT was selected due to its status as the most sophisticated model currently available in the market. For open-source, LLaMA was chosen, emanating from a renowned entity like Meta. An additional factor was the availability of complimentary credits through various web services such as TogetherAI and Replicate, eliminating the need for high-end GPU systems to operate LLaMa. The models ultimately chosen for this project include:

\begin{itemize}

    \item \textbf{Llama2 $7$B}: This model, part of Meta's LLaMA lineup, boasts $7$ billion parameters. It is adept at basic language comprehension and generation tasks, offering efficiency in processing.
    
    \item \textbf{Llama2 $13$B}: With a parameter count of $13$ billion, this variant represents a mid-tier option in the LLaMA range. It surpasses the $7$B model in handling more intricate language tasks, providing enhanced contextual understanding.

    \item \textbf{Llama2 $70$B}: The most extensive model in the LLaMA series, it encompasses $70$ billion parameters. Its considerable size and capacity render it highly proficient in complex language interpretation and generation.
    
    \item \textbf{GPT $3.5$}: A version of OpenAI's GPT-$3.5$, this model is recognized for its effective performance and optimization. Possessing $175$ billion parameters, it offers robust capabilities for complex language tasks, although not as expansive as GPT-$4$.
    
    \item \textbf{GPT $4$}: With nearly $1.8$ trillion parameters, GPT-$4$ stands as one of the most advanced AI language models to date, capable of handling both textual and visual inputs. Its immense size and multimodal function make it exceptionally suitable for tasks requiring depth and nuance.
    
\end{itemize}

\subsection{System Prompt}

System prompts serve as the initial instructions or inputs that trigger a specific task or process in an LLM. Unlike task-specific models, LLMs are not inherently fine-tuned for particular tasks. Therefore, system prompts are crucial for eliciting the desired behavior from LLMs for specific applications. Each LLM comes with a default system prompt, which must be customized for specialized tasks. These prompts, by specifying the topic, style, or unique requirements, direct the model to produce pertinent and focused responses. To minimize the occurrence of erroneous or irrelevant outputs, commonly referred to as 'hallucinations,' system prompts must be precise and straightforward. It is important to note that there is no universal system prompt suitable for all tasks; developing an effective prompt often requires several iterations and a trial-and-error approach. In this project, we implemented five distinct system prompts, employing various techniques to strike a balance between the quality and quantity of the responses generated.

\subsubsection{Few Shot Prompting}

Despite well-crafted system prompts, LLMs may occasionally struggle to adhere to the given instructions. In such instances, the technique of few-shot prompting becomes invaluable. Although LLMs are not specifically trained on task-oriented data, they possess the capability to discern patterns from minimal input. By introducing a limited yet representative set of examples, LLMs can effectively grasp the nuances of a task. For our project, we supplied the LLMs with both positive and negative examples of claim existence and scientific verifiability. This approach not only enhances the models' ability to distinguish between different types of data but also aids in refining their response accuracy, ensuring a more reliable and contextually appropriate output that aligns closely with the intended task requirements.

\subsubsection{Instruction Pattern}

The instruction pattern technique enables us to prompt Large Language Models (LLMs) to simulate roles, such as imaginary classifiers or operators. In our specific scenario, we directed the LLMs to assume the role of a "COVID-19 Tweets Classifier." This methodology involves instructing the models to categorize tweets in accordance with predefined guidelines. Adopting this approach allows the LLMs to apply a structured framework for analyzing and classifying tweet content, thereby aligning their output more closely with the specific nuances and requirements of COVID-19-related discourse. This strategy not only enhances the relevance and precision of the models' responses but also leverages their advanced language understanding capabilities for specialized content categorization.

\subsection{Configurable Parameters}

\begin{itemize}
    \item \textbf{Max New Tokens}: This parameter determines the length of the response generated by the model after receiving a prompt. A token represents a portion of text, which could be a complete word, part of a word, or punctuation. The model will continue generating tokens up to this preset limit. For instance, a max new tokens setting of $50$ restricts the response to a maximum of $50$ tokens. 
    
    \item \textbf{Temperature}: Temperature acts as a hyperparameter in Large Language Models (LLMs), influencing the degree of randomness when predicting subsequent words or phrases in the model's output. For tasks that require a high level of creativity, such as story writing, a greater temperature value is beneficial, fostering more imaginative responses. Conversely, a lower temperature setting is more appropriate for tasks that demand factual accuracy and logical reasoning, where creativity is less prioritized.

    \item \textbf{Top P (Nucleus Sampling)}: Serving as a threshold for probability, Top P influences the diversity and coherence of the generated text. It balances between randomness and restriction by not limiting choices strictly to the most likely words. 
    
    \item \textbf{Top K}: The Top K parameter plays a crucial role in the text generation process of LLMs. It limits the model's consideration to the $K$ most probable next words at each step of generating text. This constraint ensures that the model's output remains focused and relevant, as it avoids delving into less probable, and often less contextually appropriate, word choices. The selection of the Top K value is a balance between maintaining coherence and allowing for enough variability in the response. A smaller Top K value leads to more predictable text, while a larger value can introduce more diversity but may also result in less coherent outputs.

    \item \textbf{Repetition Penalty}: The Repetition Penalty parameter is a mechanism designed to enhance the quality and readability of the text generated by LLMs. By penalizing the model for selecting words or phrases it has already used, this parameter helps prevent redundant or monotonous language in the output. The degree of this penalty can be adjusted: a higher value makes the model more averse to repetition, encouraging a more varied and dynamic use of language. This feature is particularly useful in longer text generation tasks, where the risk of repetition is higher. Balancing this parameter is key to ensuring that the text remains natural and engaging, without losing coherence or straying off-topic.

\end{itemize}

\section{Experimental Setup} \label{setup}

\subsection{System Prompt Versions}

It is imperative to fine-tune system prompts until we end up with an effective version. This is why it is suggested to conduct trials and errors with multiple prompt versions. Not all prompt versions are suitable for all tasks. For this experiment, we experimented with the following $5$ different prompt versions:
\begin{itemize}
    \item \textbf{Few Shot Prompting (FSP)}: This approach incorporates a limited set of examples and counterexamples pertinent to a particular category, enhancing the LLM's focus and understanding of specific concepts.

    \begin{Verbatim}[frame=topline,label=Few Shot Prompting,framesep=3mm]
    verifiable_tweets = """
        # <3 examples of verifiable tweets>
    """

    non_verifiable_tweets = """
        # <3 examples of nonverifiable tweets>
    """

    system_message = """
    Imagine you're a COVID-19 tweets 
    classifier. 
    
    You need to determine whether tweets 
    fall into the verifiable claim category.
    
    The tweets will be delimited with 
    {delimiter} characters.
    
    {verifiable_tweets}
    
    {non_verifiable_tweets}

    If the tweet is scientifically verifiable, 
    return 1. 
    
    Otherwise, return 0.
    """
    \end{Verbatim}
    
    \item \textbf{Guidelines with Few Shot Prompting}: This variant integrates explicit guidelines alongside examples, clarifying the criteria defining a certain category, thereby aiding the LLM in more accurate categorization.

    \begin{Verbatim}[frame=topline,label=Guidelines with FSP,framesep=3mm]
    system_message = """
    # prompt start
    ...

    Use the following guidelines to make your 
    decision:
    1. Direct statements about the COVID-19 
    virus, including its origin, 
    transmission methods, prevention methods, 
    or symptoms, are scientifically verifiable.
    2. Opinionated, anecdotal, 
    or hearsay claims about COVID-19 topics 
    may be scientifically verifiable.
    3. Reports on COVID-19 cases, 
    COVID-19-related deaths, 
    or instances of someone testing positive 
    for COVID-19 is scientifically verifiable.
    4. Someone making an observation 
    is not scientifically verifiable.
    5. Instructions, information, 
    notifications, or announcements 
    about COVID-19 topics that do not include 
    opinions are not scientifically verifiable.

    # rest of the prompt
    ...
    """
    \end{Verbatim}

    \item \textbf{Guidelines with Few Shot Prompting and Emotional Stimuli}: Drawing on insights from \cite{li2023large}, this method employs emotionally charged sentences or phrases, enhancing the LLM's engagement and performance. The addition of emotional stimuli aims to leverage the model's response sensitivity for more nuanced task execution.

    \begin{Verbatim}[frame=topline,label=Guidelines + FSP + Emotional Stimuli,framesep=3mm]
    system_message = """
    # prompt start

    ...

    This task is very important to my career. 
    You'd better be sure. 
    Make sure to take another look at your 
    response before responding.
    """
    \end{Verbatim}
    
    \item \textbf{Chain of Thought}: A widely recognized technique in the machine learning domain, this strategy involves guiding the LLM to methodically ponder the problem. Particularly effective in logic-based, mathematical, or intricate decision-making tasks, it encourages the model to sequentially process the problem, thus yielding more precise and interpretable responses by avoiding hasty conclusions or logical jumps.

    \begin{Verbatim}[frame=topline,label=Chain of Thought,framesep=3mm]
    system_message = """
    # prompt start
    ...
    
    You should employ the following 
    chain-of-thought process examples 
    for the classification:

    1. Tweet: "COVID-19 is transmitted 
    through respiratory droplets."
    Thought Process: This is a direct 
    statement about how COVID-19 is 
    transmitted. 
    It can be scientifically verified 
    through studies and data.
    Classification: Scientifically Verifiable

    2. Tweet: "COVID-19 has led to more 
    people working from home."
    Thought Process: This statement is about 
    the impact on work habits, a 
    business-related issue, not a scientific 
    claim about the virus.
    Classification: Not Scientifically 
    Verifiable

    # prompt end
    ...
    """
    \end{Verbatim}
    
    \item \textbf{Clue and Reasoning Enhanced Prompting (CARP)}: Introduced by \cite{sun2023text}, CARP aims to structure the prompting process more systematically. It's tailored to bolster the LLM's capacity for logical reasoning, utilizing clues embedded in the prompt to guide response generation in a more reasoned and coherent manner.

    \begin{Verbatim}[frame=topline,label=CARP,framesep=3mm]
    system_message = """
    # prompt start
    ...
    
    Apply the following steps:

    1. CLUE IDENTIFICATION: Determine the 
    presence of direct statements, reports, 
    or factual claims related to COVID-19 
    using keywords and context within the 
    Tweet.

    2. REASONING PROCESS: Analyze the clues 
    to ascertain if they align with the 
    scientific facts, data, or reputable 
    health authority guidelines.

    3. VERIFICATION DETERMINATION: Decide if 
    the tweet's claim is scientifically 
    verifiable, based on the evidence 
    and reasoning.
    
    Example 1:
    TWEET: "New research indicates that 
    COVID-19 can remain on surfaces for 
    days."
    CLUES: "New research," "COVID-19," 
    "remain on surfaces," "days"
    REASONING: The claim is presented as a 
    finding from new research, which is a 
    direct statement about the virus's 
    transmission and is likely based on 
    scientific studies.
    VERIFICATION: Scientifically Verifiable

    Example 2:
    TWEET: "The government's response to 
    COVID-19 will surely boost the economy."
    CLUES: "Government's response," "COVID-19," 
    "boost," "economy."
    REASONING: The input is making a 
    speculative assertion about the impact 
    of the government's COVID-19 response 
    on the economy. While it relates to 
    COVID-19, it is framed as a prediction 
    rather than a fact and does not directly 
    pertain to the scientific aspects of the 
    virus itself. The claim is about 
    economic impact, which is outside the 
    scope of scientific verification.
    
    VERIFICATION: Not Scientifically Verifiable

    # prompt end
    ...
    """
    \end{Verbatim}

\end{itemize}

\subsection{System Prompt Format}

For Llama 2, the prompts are structured with specialized tokens like $<s>$, $<INST>$, and $<SYS>$, each serving a specific purpose in guiding the model's response. These tokens delineate the start of a sequence, provide instructions, and differentiate between system-generated content and user input, respectively.

    \begin{Verbatim}[frame=topline,label=LLaMA Example Prompt Format,framesep=3mm]
    <s>[INST] <<SYS>>
    You are a helpful, and honest assistant. 
    Always answer as helpfully as possible, 
    while being safe. 
    
    If you don't know the answer to a question, 
    please don't share false information.
    <</SYS>>
    
    There's a llama in my garden. 
    What should I do? [/INST]
    \end{Verbatim}

In contrast, GPT employs a more conversational format, often using a series of messages in a dialogue structure. Each message in ChatGPT is typically formatted as a dictionary with keys indicating the "role" (system or user) and "content" of the message. This approach aligns with GPT's design for more natural, conversational interactions.

\begin{Verbatim}[frame=topline,label=GPT Example Prompt Format,framesep=3mm]
    System Message: "You are a helpful, 
    and honest assistant. Always answer 
    as helpfully as possible, while being safe."
    
    User Message: "There's a llama in my garden. 
    What should I do?"
    
    AI Response: If there's a llama in your garden, 
    here are some steps you can take: ...
    \end{Verbatim}

\subsection{Parameter Values}

\begin{itemize}
    \item \textbf{Max New Tokens}: In our experiment, we set this limit to $1500$ tokens, aiming for responses that are comprehensive yet succinct, including the model's reasoning process.
    \item \textbf{Temperature}: The temperature value was set at $0.2$, ensuring that the LLM models' responses stayed consistent across all tweets.
    \item \textbf{Top P}: For our purposes, a Top P value of $0.4$ was chosen, aiming to achieve a blend of diversity and coherence in the text generation.
    \item \textbf{Top K}: A Top K value of $50$ was selected, which helps maintain textual coherence by avoiding highly improbable word choices, while still allowing for a degree of variability in the output.
    \item \textbf{Repetition Penalty}:  We opted for a value of $1$ for the Repetition Penalty, treating all words equally without imposing penalties for reuse. This neutral setting avoids artificially influencing the model's natural language generation.
\end{itemize}

\subsection{Local GPU Cluster Setup}

The Llama models are available as open-source, with their weights and biases accessible for download under a lenient license. However, to execute these models locally, substantial GPU resources are required. The base model, Llama 2 7B, for instance, demands a minimum of 30GB of GPU memory for full precision operation. Running the larger Llama 2 70B model was infeasible in our setup, as it requires at least 140GB of memory, even when operating in half-precision mode. The bash script provided below details the specific parameters used to deploy the Llama 2 models on our GPU cluster.

\begin{Verbatim}[frame=topline,label=GPU Cluster Setup,framesep=3mm]
    #!/bin/bash
    #SBATCH --job-name="LLM-Claim" 
    #SBATCH --partition=peregrine-gpu
    #SBATCH --nodes=1                               
    #SBATCH --ntasks=1                              
    #SBATCH --cpus-per-task=1                       
    #SBATCH --mem=80G
    #SBATCH --gres=gpu:a100-sxm4-80gb:1
    #SBATCH --time=08:00:00                         
    #SBATCH --output=my-job.out # log file
    #SBATCH --error=my-job.err  # error file
\end{Verbatim}

\subsection{Cloud AI Providers}

To utilize Llama models without the constraints of high GPU resource demands, cloud AI providers like Replicate and TogetherAI offer a more flexible and convenient solution. These platforms enable users to quickly start working with the models, eliminating the need for intricate GPU configurations. Additionally, these services often provide complimentary credits upon signup, further easing the initial setup process. Utilizing their services is straightforward, typically involving a simple API call using an API key. These providers also offer a playground feature, allowing users to experiment with a variety of models.

However, there are some limitations to consider. These cloud platforms may occasionally face outages, and the free tier usage often comes with rate limiting, which can restrict the frequency of API calls. Despite these potential drawbacks, turning to cloud AI providers remains a valuable option, particularly for those who might not have access to complex GPU setups. This approach allows for easier and more accessible use of advanced AI models like Llama.

\section{Evaluation Metrics} \label{evaluation}

In the evaluation process, four well-known metrics were utilized: Accuracy, Precision, Recall, and F1 Score.

\begin{itemize}
    \item \textbf{Accuracy}: Accuracy measures the proportion of overall correct predictions made by the model. It is calculated as the sum of true positives (TP) and true negatives (TN) divided by the total number of predictions, including both correct and incorrect predictions (TP, TN, false positives (FP), and false negatives (FN). Mathematically, it is represented as \[ \text{Accuracy (A)} = \frac{TP + TN}{TP + TN + FP + FN} \]

    \item \textbf{Precision}: Precision evaluates the accuracy of the model's positive predictions. It is defined as the ratio of true positives to the total number of positive predictions made by the model, which includes both true positives and false positives. The formula is \[ \text{Precision (P)} = \frac{TP}{TP + FP} \] This metric is crucial in contexts where the cost of a false positive is significant. Within the realm of identifying scientific claims, a lower precision indicates a tendency for the model to erroneously label factual information as misinformation. Such an outcome is undesirable for the intended functionality of the model, necessitating the achievement of higher precision values to ensure accurate categorization.

    \item \textbf{Recall}: Recall assesses the model's ability to correctly identify actual positive instances. It is calculated as the ratio of true positives to the sum of true positives and false negatives represented as \[ \text{Recall (R)} = \frac{TP}{TP + FN} \]. This metric is particularly important in scenarios where missing a positive instance has serious implications. In the process of detecting and classifying scientific claims, a lower recall value signifies that the model is incorrectly identifying misinformation as factual. This issue carries more severe consequences than having a lower precision and must be rigorously avoided to ensure the integrity and reliability of the classification process.

    \item \textbf{F1 Score}: F1 Score is the harmonic mean of Precision and Recall, striking a balance between them. It is especially useful when seeking a single metric to reflect both the precision and recall of the model. The F1 Score is calculated as \[ \text{F1 Score (F1)} = 2 \times \frac{P \times R}{P + R} \] This metric is beneficial for comparing models where a trade-off between precision and recall exists.
\end{itemize}

\section{Results} \label{results}

The tables below present the performance metrics of different models on two tasks: Claim Existence and Verifiability.

\textbf{Llama 2 $7$B} (Table \ref{llama2-7b}): In the Claim Existence task, this model achieved an F1-score of 0.54, indicating a balanced performance in precision and recall. For the Verifiable task, it showed a higher F1-score of 0.72, suggesting better effectiveness in identifying verifiable claims.

\begin{table}[h]
\setlength{\abovecaptionskip}{-10pt} 
\setlength{\belowcaptionskip}{10pt} 
\resizebox{\columnwidth}{!}{
\scriptsize 
\begin{tabular}{|c|c|c|c|c|}
\hline
\textbf{Task}        & \textbf{P} & \textbf{R}  & \textbf{A} & \textbf{F1} \\ \hline
Claim Existence    & 0.53  & 0.55   & 0.54  & 0.54   \\ \hline
Verifiable             & 0.86  & 0.63   & 0.60  & 0.0.72   \\ \hline
\end{tabular}
}
\newline
\caption{Results for Llama 2 7B}
\label{llama2-7b}
\end{table}

\textbf{Llama 2 $13$B} (Table \ref{llama2-13b}): The performance of this model was slightly lower than Llama 2 7B in both tasks, with an F1-score of 0.52 in Claim Existence and 0.71 in Verifiable tasks. It demonstrates a consistent but somewhat reduced efficiency compared to Llama 2 7B.

\begin{table}[h]
\setlength{\abovecaptionskip}{-10pt} 
\setlength{\belowcaptionskip}{10pt} 
\resizebox{\columnwidth}{!}{%
\scriptsize 
\begin{tabular}{|c|c|c|c|c|}
\hline
\textbf{Task}        & \textbf{P} & \textbf{R}  & \textbf{A} & \textbf{F1} \\ \hline
Claim Existence   & 0.52  & 0.52   & 0.51  & 0.52   \\ \hline
Verifiable             & 0.81  & 0.63   & 0.60  & 0.71   \\ \hline
\end{tabular}
}
\newline
\caption{Results for Llama 2 13B}
\label{llama2-13b}
\end{table}

\textbf{Llama 2 $70$B} (Table \ref{llama2-70b}): Surprisingly, Llama 2 70B underperformed in the Verifiable task with an F1-score of 0.63, lower than its smaller counterparts. In the Claim Existence task, it achieved a similar F1-score of 0.52, consistent with Llama 2 13B.

\begin{table}[h]
\setlength{\abovecaptionskip}{-10pt} 
\setlength{\belowcaptionskip}{10pt}
\resizebox{\columnwidth}{!}{%
\scriptsize 
\begin{tabular}{|c|c|c|c|c|}
\hline
\textbf{Task}        & \textbf{P} & \textbf{R}  & \textbf{A} & \textbf{F1} \\ \hline
Claim Existence    & 0.53  & 0.52   & 0.52  & 0.52   \\ \hline
Verifiable             & 0.61  & 0.64   & 0.56  & 0.63   \\ \hline
\end{tabular}
}
\newline
\caption{Results for Llama 2 70B}
\label{llama2-70b}
\end{table}

\textbf{GPT $3.5$} (Table \ref{gpt3-turbo}): This model exhibited a notable improvement in the Claim Existence task with an F1-score of 0.61 and a significantly better performance in the Verifiable task with an F1-score of 0.87. It outperformed the Llama 2 models in detecting verifiable claims.

\begin{table}[h]
\setlength{\abovecaptionskip}{-10pt} 
\setlength{\belowcaptionskip}{10pt}
\resizebox{\columnwidth}{!}{%
\scriptsize 
\begin{tabular}{|c|c|c|c|c|}
\hline
\textbf{Task}        & \textbf{P} & \textbf{R}  & \textbf{A} & \textbf{F1} \\ \hline
Claim Existence    & 0.73  & 0.52   & 0.53  & 0.61   \\ \hline
Verifiable             & 0.77  & 0.66   & 0.59  & 0.70   \\ \hline
\end{tabular}
}
\newline
\caption{Results for GPT 3.5 Turbo}
\label{gpt3-turbo}
\end{table}

\textbf{GPT $4$} (Table \ref{gpt4}): GPT 4 outshone all other models, registering the highest F1-scores of 0.65 in Claim Existence and 0.76 in Verifiable tasks. This underscores its superior capability in both identifying the existence of claims and assessing their verifiability. 

\begin{table}[h]
\setlength{\abovecaptionskip}{-10pt} 
\setlength{\belowcaptionskip}{10pt}
\resizebox{\columnwidth}{!}{%
\scriptsize 
\begin{tabular}{|c|c|c|c|c|}
\hline
\textbf{Task}        & \textbf{P} & \textbf{R}  & \textbf{A} & \textbf{F1} \\ \hline
Claim Existence    & 0.79  & 0.53   & 0.55  & \textbf{\underline{0.65}}   \\ \hline
Verifiable             & 0.87  & 0.64   & 0.62  & \textbf{\underline{0.76}}   \\ \hline
\end{tabular}
}
\newline
\caption{Results for GPT 4}
\label{gpt4}
\end{table}

\section{Result Analysis} \label{result-analysis}

GPT 4 outperformed all other models, likely attributable to its advanced architecture and comprehensive training. Within the Llama 2 series, the 7b model demonstrated a marginal superiority over the 13b, while, intriguingly, the 70b model exhibited the least effective performance. This variation in performance could be linked to differences in the training methodologies of these models. Furthermore, certain features of the tweets, such as colloquial language or ambiguous phrasing, might have affected the models' classification accuracy.

Across all models, the observed lower recall values suggest a tendency to overlook true positives. This could stem from the models' limitations in recognizing subtle details or potential issues with how the dataset was annotated. While the high precision of these models minimizes the likelihood of false positives, thereby reducing false alarms, the lower recall raises concerns about the possibility of missing instances of misinformation. 

In enhancing these models, a focus on improving recall without compromising precision could be vital, especially in tasks where detecting every instance of misinformation is crucial for maintaining the integrity and reliability of the information being disseminated.

\section{Future Work} \label{future-work}

Future research directions are inspired by the paper on Retrieval Augmented Generation (RAG) \cite{rag} and other methodologies:

\begin{itemize}
    \item \textbf{Retrieval Augmented Generation (RAG)}: While LLMs fall short in knowledge-intensive tasks compared to task-specific architectures, RAG offers a promising solution. It involves augmenting the input query with information retrieved from a Knowledge Base, using vector databases for numerical vector representation and similarity matching techniques like Cosine Similarity or TF-IDF.
    \item \textbf{Fine-Tuning on Specific Datasets}: To better capture the intricacies and specific requirements of a dataset, feeding it directly into LLMs for fine-tuning is recommended. Detailed guidance on this process can be found at OpenAI’s fine-tuning platform.
    \item \textbf{Experimentation with Varied Prompt Techniques}: Exploring different and innovative prompting techniques could yield improvements in the model's performance.

    \item \textbf{Hybrid Approaches (LLMs + SLMs)}: Combining the strengths of Large Language Models (LLMs) and Specific Language Models (SLMs) could lead to a more robust and effective system, particularly in dealing with nuanced and complex tasks.
\end{itemize}

These future research avenues aim to address the current limitations and explore innovative ways to enhance the performance and applicability of language models in various real-world scenarios.

\section{Conclusion} \label{conclusion}

This project embarked on a comprehensive exploration of the capabilities of Large Language Models (LLMs) in the context of detecting and classifying scientific claims related to COVID-19. Our investigation revealed insightful findings about the performance of various LLMs, including different configurations of the Llama 2 series and the advanced GPT models. GPT 4 emerged as the standout performer, demonstrating superior efficacy in identifying and verifying scientific claims, attributed to its sophisticated architecture and extensive training.

The journey revealed the critical importance of system prompts in guiding the models' analysis, highlighting the nuanced balance between art and science in their creation. Moreover, the subtlety of language inherent in tweets presented a notable challenge, with the LLMs sometimes struggling to fully grasp the author's intended tone or nuances—a key aspect in accurate context understanding. Despite the challenges, the project's success in utilizing LLMs for complex, real-world tasks like misinformation detection has been encouraging.

Looking forward, the potential of Retrieval Augmented Generation (RAG) and fine-tuning models on specific datasets presents exciting avenues for further enhancing the effectiveness of LLMs. The exploration of varied prompt techniques and hybrid approaches combining LLMs with Specific Language Models (SLMs) offers promising paths to overcome existing limitations.

In conclusion, this project not only highlighted the capabilities and limitations of current LLMs in handling complex, real-world tasks but also opened up pathways for future research and development in the field. The insights gained from this study are a stepping stone towards more nuanced and effective use of AI in the realm of information verification and dissemination, especially in critical areas like public health communication during global pandemics. The findings underscore the importance of addressing the nuanced language challenges in tweets to improve the reliability and accuracy of LLMs in practical applications.



\bibliographystyle{ACM-Reference-Format}
\bibliography{sample-base}


\end{document}